\documentclass[letterpaper, 10 pt, conference]{ieeeconf}

\IEEEoverridecommandlockouts                              

\usepackage{graphicx} 
\usepackage{booktabs}
\usepackage{multirow}
\usepackage{url} 
\usepackage{subcaption}
\usepackage[table]{xcolor}

\title{\LARGE \bf
AD4AD: Benchmarking Visual Anomaly Detection Models\\ for Safer Autonomous Driving
}

\author{Fabrizio Genilotti$^{1*}$, Arianna Stropeni$^{1*}$, Gionata Grotto$^{*}$,\\ Francesco Borsatti$^{1}$, Manuel Barusco$^{1}$, Davide Dalle Pezze$^{1}$ and Gian Antonio Susto$^{1}$%
\thanks{*These authors contributed equally to this work}%
\thanks{$^{1}$The authors are with the University of Padova, Italy.
        { \{fabrizio.genilotti, arianna.stropeni, gionata.grotto\}@studenti.unipd.it,
        \{francesco.borsatti, manuel.barusco\}@phd.unipd.it,
        \{davide.dallepezze, gianantonio.susto\}@unipd.it}}%
}

\begin{document}

\maketitle
\thispagestyle{empty}
\pagestyle{empty}

\begin{abstract}
The reliability of a machine vision system for autonomous driving depends heavily on its training data distribution. When a vehicle encounters significantly different conditions, such as atypical obstacles, its perceptual capabilities can degrade substantially. Unlike many domains where errors carry limited consequences, failures in autonomous driving translate directly into physical risk for passengers, pedestrians, and other road users.
To address this challenge, we explore Visual Anomaly Detection (VAD) as a solution.
VAD enables the identification of anomalous objects not present during training, allowing the system to alert the driver when an unfamiliar situation is detected. Crucially, VAD models produce pixel-level anomaly maps that can guide driver attention to specific regions of concern without requiring any prior assumptions about the nature or form of the hazard.
We benchmark eight state-of-the-art VAD methods on AnoVox, the largest synthetic dataset for anomaly detection in autonomous driving.
In particular, we evaluate performance across four backbone architectures spanning from large networks to lightweight ones such as MobileNet and DeiT-Tiny. 
Our results demonstrate that VAD transfers effectively to road scenes. 
Notably, Tiny-Dinomaly achieves the best accuracy–efficiency trade-off for edge deployment, matching full-scale localization performance at a fraction of the memory cost.
This study represents a concrete step toward safer, more responsible deployment of autonomous vehicles, ultimately improving protection for passengers, pedestrians, and all road users.
\end{abstract}

\section{Introduction}
Autonomous driving systems rely on machine learning models trained on large datasets to perceive and interpret the surrounding environment. However, the reliability of such systems is fundamentally bounded by their training data distribution. When a vehicle encounters conditions that deviate significantly from what was seen during training, e.g. atypical obstacles, unusual road configurations, or rare events, its perceptual capabilities can degrade substantially. 
\\
Unlike many domains where prediction errors carry limited consequences, failures in autonomous driving translate directly into physical risk for passengers, pedestrians, and other road users. A system must therefore not only perform well under expected conditions, but also handle edge cases correctly, or recognize its own limitations when encountering them.
\\
A natural response to this challenge is to equip the system with the ability to detect when something unknown is present in the scene. Visual Anomaly Detection (VAD) can offer precisely this capability: VAD models are trained exclusively on normal, expected data and, at inference time, identify anything that deviates from the learned distribution without requiring labeled anomalous examples and without presupposing what form the anomaly may take. This open-world property is critical in the autonomous driving context, where the space of possible road hazards is virtually unbounded and cannot be exhaustively anticipated at training time.
\\
In addition,  VAD models produce pixel-level anomaly maps that spatially localize the detected deviation within the scene. This is particularly valuable for driver-assistance applications: rather than issuing a generic alert, the system can direct the driver's attention to the specific region of the scene where the anomaly is located, effectively reducing reaction time and supporting faster, more informed decision-making.
\\
Despite the growing maturity of VAD in industrial inspection and medical imaging, its application to autonomous driving remains unexplored. Existing benchmarks and evaluation protocols are not designed for road scenarios, and it remains unclear whether state-of-the-art VAD methods generalize effectively to the visual characteristics of on-road environments.
\\
Our contributions can be summarized as follows:
\begin{itemize}

    \item We present the \textbf{first systematic evaluation of Visual Anomaly Detection in the autonomous driving domain}, benchmarking eight state-of-the-art VAD methods on AnoVox, a dataset specifically designed for anomaly detection on the road.

    \item We perform an \textbf{edge-oriented analysis}, by comparing the use of feature extractors of varying size and nature, including CNN-based backbones and lightweight Vision Transformers, providing concrete guidance on the performance–efficiency trade-off for onboard deployment in resource-constrained automotive hardware.

    \item The results obtained on AnoVox demonstrate that VAD models are effective at this task, and that the anomaly maps provide meaningful spatial guidance for driver alerting,  providing strong baselines for future work in the field.

\end{itemize}
The remainder of this paper is organized as follows: section \ref{sec:related_work} reviews related work on VAD in general and its applications to autonomous driving; section \ref{sec:methodology} describes the dataset, the evaluated models and the chosen feature extractors; section \ref{sec:results} presents and discusses the results; finally, Section \ref{sec:conclusions} draws conclusions and outlines future directions.

\begin{figure}[!th]
    \centering
    \includegraphics[width=\columnwidth]{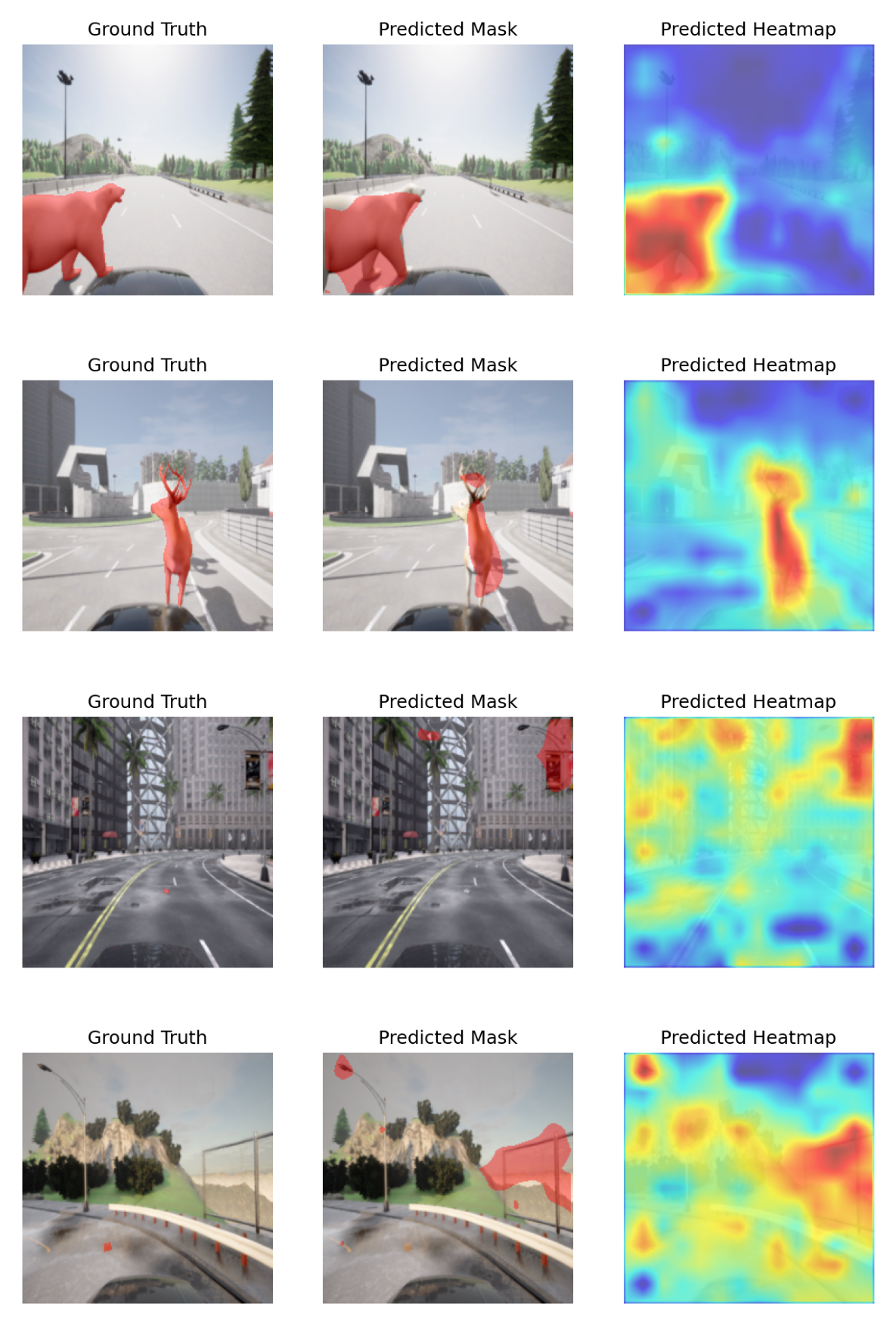}
    \caption{Anomaly maps produced by Dinomaly using a DeiT-Small backbone. We report two cases where the model performs well and two failure cases.}
    \label{fig:heatmaps}
\end{figure}

\section{Related Work}
\label{sec:related_work}

\subsection{Visual Anomaly Detection}

Visual Anomaly Detection (VAD) plays a crucial role in many computer vision applications, including manufacturing and healthcare. VAD models offer two main advantages: first, they are typically trained in an unsupervised manner using only anomaly-free samples, bypassing the costly process of collecting and labeling large numbers of anomalous samples; second, they produce pixel-level anomaly maps that enhance interpretability, operator decision-making, and end-user trust.
State-of-the-art VAD models generally fall into two main categories: reconstruction-based methods and feature embedding–based methods.
\\
\textbf{Reconstruction-based methods} employ generative models to learn the distribution of normal data, identifying anomalies at inference time through high reconstruction errors. Common approaches include Autoencoders, GANs, and Diffusion Models \cite{draem, diffusionad}. However, these methods are often computationally expensive, which can limit their applicability in real-time or resource-constrained scenarios. 
\\
\textbf{Feature embedding–based methods}, in contrast, exploit representations extracted from pretrained neural networks, avoiding explicit image reconstruction and achieving higher computational efficiency.
This family is further organized into three subcategories: \textbf{Teacher–Student methods}, which detect anomalies through discrepancies between teacher and student feature maps (e.g. STFPM \cite{st_pyramid}); \textbf{Memory Bank methods}, which store normal feature representations for comparison at inference time (e.g., PaDiM \cite{PaDiM}, PatchCore \cite{patch}, CFA \cite{lee2022cfa}); and \textbf{Normalizing Flow methods}, which map data distributions to a normal distribution for likelihood-based anomaly detection (e.g. FastFlow \cite{yu2021fastflow}).

\subsection{Anomaly Detection for Autonomous Driving}
As anticipated, the detection of hazards that may compromise the safety of road users is crucial.
To tackle this problem, numerous datasets and benchmarks have been presented in literature as discussed in \cite{Bogdoll_2023_adbenchmarksurvey}, each characterized by different properties, such as their synthetic or real-world nature and their underlying definition of normality. 
\\
\cite{chan2021segmentmeifyoucanbenchmarkanomalysegmentation} introduces two real-world datasets: RoadAnomaly21, containing internet-collected road images with anomalous objects, and RoadObstacle21, recorded in a limited set of street environments. Authors of \cite{hendrycks2022scalingoutofdistributiondetectionrealworld} propose StreetHazards, a synthetic CARLA-based dataset featuring 250 anomalies across multiple town scenes and weather conditions, and BDD-Anomaly, derived from BDD100K by treating motorcycles, trains, and bicycles as anomalous classes at test time. Lost and Found \cite{lostandfound} targets small hazard detection across diverse real road scenarios, covering 42 object types.
Fishyscapes \cite{fishyscapes} is an evaluation benchmark providing two validation sets: FS Lost and Found and FS Static, the latter obtained by overlaying anomalous objects onto Cityscapes images.
\\
However, these datasets present limitations in the context of Anomaly Detection for Autonomous Driving (AD4AD): Lost and Found lacks road diversity; Fishyscapes relies on artificial anomalies; StreetHazards includes unrealistic anomalies in irrelevant driving regions; BDD-Anomaly defines anomalies over an unsuitable class set; and both RoadAnomaly21 and RoadObstacle21 contain only anomalous images. 
\\
While these datasets present limitations, the synthetic AnoVox benchmark \cite{bogdoll2024anovox}, which contains various scenarios, multimodal sensor data and images with more realistic anomalies positioned in relevant road regions, represents a more suitable choice to evaluate AD4AD.
\\
\\
The majority of camera-based anomaly detection methods proposed in the AD4AD literature are built around a pretrained closed-set semantic segmentation network, typically trained in a fully supervised manner on the Cityscapes benchmark, which defines a fixed taxonomy of 19 known classes \cite{bogdoll2022anomaly}. These methods treat the segmentation network itself as a proxy for normality.
A separate line of work adopts reconstruction-based approaches in which an autoencoder is trained to reproduce normal-appearing inputs, with high reconstruction error serving as the anomaly signal. However, as is well established in the broader Visual Anomaly Detection literature, autoencoders tend to generalize beyond their training distribution and reconstruct anomalous regions with surprisingly low error, undermining their reliability as anomaly detectors. 
\\
VAD methods learn the distribution of normal visual appearances directly from unlabeled normal samples. At inference time, any region deviating sufficiently from the learned distribution is flagged as anomalous.
This open-world property is critical in autonomous driving, where the space of possible road hazards is virtually unbounded and cannot be exhaustively anticipated at training time. Additionally, VAD methods produce pixel-level anomaly maps that spatially localize detected deviations, enabling driver assistance systems to direct attention to specific regions rather than issuing generic alerts.

\section{Methodology}
\label{sec:methodology}

\subsection{AnoVox}

A fundamental requirement for model evaluation is a benchmark that offers both a rigorous, well-defined notion of normality and a wide range of challenging, diverse anomalous scenarios. Among the existing benchmarks, we adopt AnoVox, currently the largest dataset for anomaly detection in autonomous driving. 
AnoVox is a synthetic benchmark based on the popular open-source CARLA simulator, which guarantees full controllability of the training and evaluation environments. In particular, AnoVox provides a simple way to distinguish normality, allowing easy evaluation of model capabilities. 
\\ 
During this work, we focus on the content anomalies defined by AnoVox, namely entities that are placed at critical points of the road and that could compromise the safety of road users. 
We used the dataset version provided at \footnote{\url{https://zenodo.org/records/8171712}}, composed of 1850 frames across 10 road scenes, as a first challenging testbed for evaluating AD4AD systems.
Given its rigorous organization and that approximately 14.8\% of the camera frames within the dataset contain anomalies, this benchmark is perfectly suitable for evaluating our methods in a standard VAD setup.
\\
While the dataset includes both RGB camera images and LiDAR point clouds, in this work, we focus on content anomalies captured using only the RGB camera, as our evaluation targets feature-based VAD methods that operate on single images. 
Since the anomalies are rare by definition, we adopt the standard evaluation protocol for VAD: the models are trained only on normal samples, and only in the test are the anomalies present. Despite the synthetic setting, AnoVox represents a significant step forward with respect to prior benchmarks in autonomous driving anomaly detection, which were generally limited in scale.

\subsection{VAD Methods}
\label{subsec:vad_methods}

As discussed in Section \ref{sec:related_work}, VAD methods fall into two main categories: reconstruction-based approaches using image-to-image generative models and feature-based approaches exploiting representations from pre-trained networks.
In the autonomous driving scenario, computation is often performed at the edge, thus requiring lightweight, efficient architectures. For this reason, we mainly adopt feature-based approaches, which can be readily adapted to edge deployments at the cost of a minimal performance loss \cite{barusco2024paste}.
In the following, we briefly describe the eight VAD methods considered in this study:
\begin{itemize}
    \item \textbf{Patchcore}: it builds a compact memory bank of representative normal patches and flags anomalies based on the distance of test patches to their nearest neighbors \cite{patch}.
    \item \textbf{Padim}: it models each spatial feature location with a multivariate Gaussian and uses the Mahalanobis distance to detect deviations as anomalies \cite{PaDiM}.
    \item \textbf{CFA}: it creates a memory of normal patch embeddings and adapts them into coupled hyperspheres to amplify the separation between normal and abnormal feature representations \cite{lee2022cfa}.
    \item \textbf{STFPM}: it is based on two networks (teacher and student) with knowledge distillation, where student and teacher feature map deviations indicate anomalies \cite{st_pyramid}.
    \item \textbf{RD4AD}: it improves the STFPM approach by considering an autoencoder-like approach where the teacher is the encoder and the student the decoder \cite{rd4ad}.
    \item \textbf{SuperSimpleNet (SSNet)}: it uses a feature adaptor like CFA and considers generating synthetic anomalies at the feature level to improve performance  \cite{rolih2025supersimplenet}.
    \item \textbf{FastFlow}: it is based on normalizing flow models to transform complex input data distributions into normal distributions, leveraging probability as a measure of normality \cite{yu2021fastflow}. 
    \item \textbf{Dinomaly}: like RD4AD, it employs an encoder-decoder architecture. 
    However, Dinomaly uses a pure Transformer encoder-decoder architecture, with a frozen, pretrained ViT serving as the encoder. The decoder then learns to reconstruct these encoder features exclusively from normal samples and to identify anomalies by exploiting discrepancies between encoder and decoder representations across multiple scales \cite{guo2025dinomaly}.
\end{itemize}

\subsection{Backbone Selection for Edge Deployment}
The choice of the feature extractor plays a central role in our scenario, since it determines the trade-off between representation richness, which correlates with detection and localization performance, and computational costs. In this work, we evaluate VAD methods across a variety of backbone architectures, covering a spectrum from  convolutional networks to Vision Transformers.
\\
Among CNN-based backbones, we focus on WideResNet-50 and MobileNet-V2, hereafter referred to as \textbf{WideResNet} and \textbf{MobileNet}, respectively. WideResNet is widely employed across feature-based approaches, thanks to the richness of the extracted feature representations. However, its complex architecture makes it unsuitable for deployment in resource-constrained environments such as onboard automotive hardware. Due to this reason, we consider MobileNet as an alternative: its architecture is designed for efficiency, and its employment in VAD has already been validated by other works \cite{barusco2024paste}. 
\\
Vision Transformer-based backbones represent a more recent paradigm in visual feature extraction. Pretrained via supervised or self-supervised objectives on large image corpora, these models produce patch-level token representations that capture long-range spatial dependencies, a property that proves advantageous for anomaly localization in complex road scenes. 
Among ViT-based backbones, we consider two variants from the DeiT family \cite{touvron2021training}: we select \textbf{DeiT-Small} and \textbf{DeiT-Tiny} as our ViT-based backbones.
DeiT-Small is chosen to provide a fair comparison with WideResNet, as the two architectures are comparable in total parameter count. DeiT-Tiny, on the other hand, serves as a lightweight alternative whose capacity is more closely aligned with compact architectures such as MobileNet.
\\
Note that, while testing the majority of our methods with all backbones, we consider only ViT-based backbones for Dinomaly, due to the nature of the method, which was specifically designed around Vision Transformers architectures.
In particular, the variant of Dinomaly employing DeiT-Tiny as its backbone is referred to as \textbf{Tiny-Dinomaly}, as this specific configuration was previously proposed for edge-oriented VAD \cite{barusco2026continual}.

\subsection{Evaluation Metrics}
\label{subsec:evaluation_metrics}

We evaluate the methods using standard metrics for visual anomaly detection:

\begin{itemize}
\item \textbf{Image-level AUROC (I-ROC)}: Area under the Receiver Operating Characteristic Curve for classifying images as normal or anomalous.
\item \textbf{Pixel-level AUROC (P-ROC)}: Area under the Receiver Operating Characteristic Curve for classifying each pixel as normal or anomalous.
\item \textbf{Image-level PR AUROC (I-AP)} and \textbf{Pixel-level PR AUROC (P-AP)}: Area under the Precision-Recall Curve summarizes the trade-off between precision and recall, and image- and pixel-level respectively.
\item \textbf{Image-level F1 Score (I-F1)} and \textbf{Pixel-level F1 Score (P-F1)}: harmonic mean of precision and recall, that evaluates whether each image is correctly classified and, at pixel level, the overlap between the predicted and ground-truth anomaly masks. Following typical approaches in VAD, we compute the F1 score at the optimal threshold that maximizes the performance on the test set.
\item \textbf{Per-Region Overlap (PRO)}: it computes the mean overlap between the predicted anomaly map and each individual ground-truth connected component, weighting all anomalous regions equally regardless of their size. This makes it a more informative metric when anomalies vary significantly in scale.
\item \textbf{Memory Footprint}: Total memory required to store model parameters and auxiliary data structures (memory bank, statistics, etc.).
\item \textbf{Inference Time}: Average time required to process a single test image on a target edge device. Inference times were collected on an Intel i5-based development system, used as a proxy for the target automotive hardware.

\end{itemize}

\begin{figure}[!thbp]
    \centering
    \includegraphics[width=\linewidth]{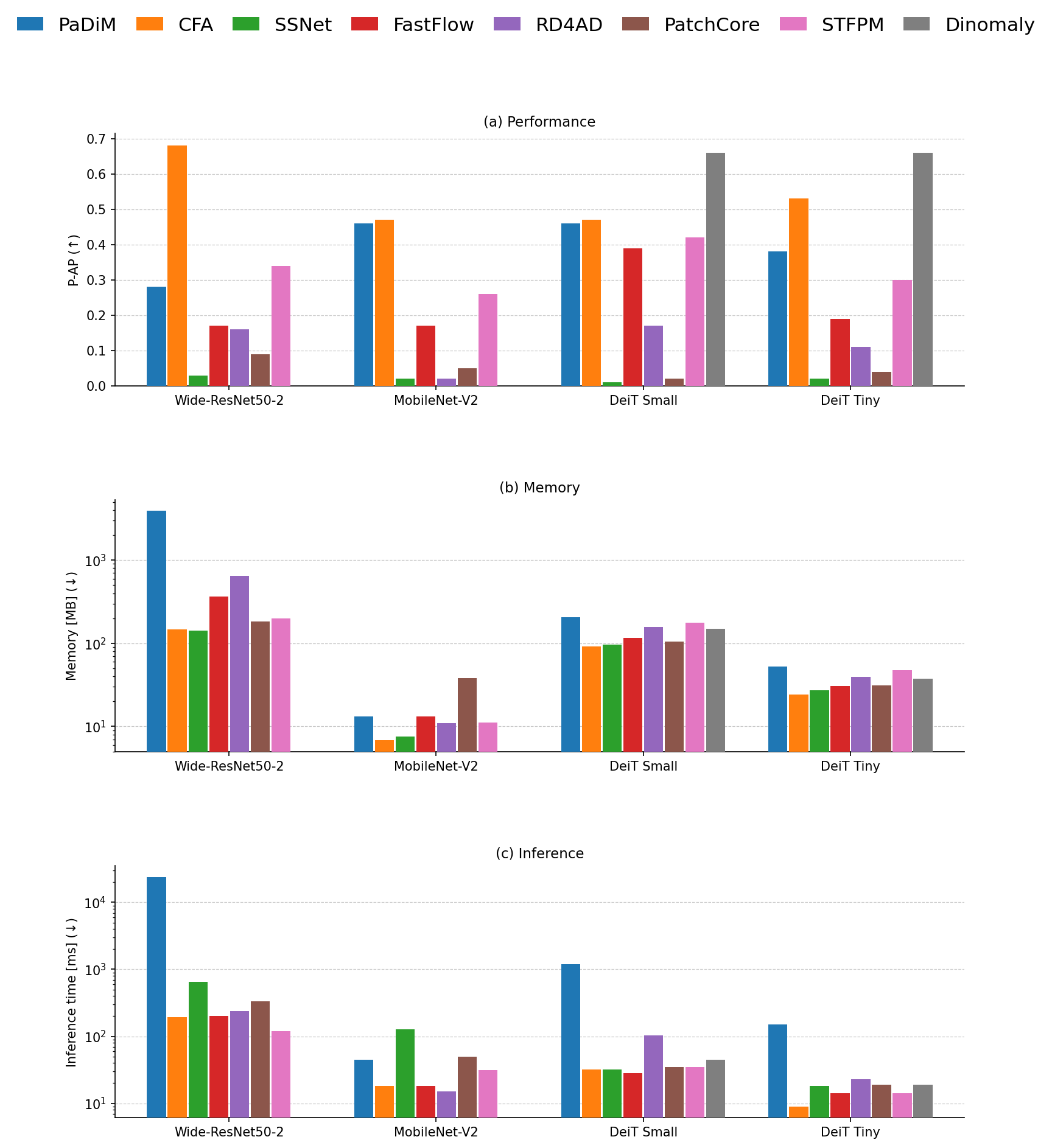}
    \caption{Comparison of VAD models for performance (P-AP), memory, and inference.}
    \label{fig:results-plot}
\end{figure}

\begin{table*}[!th]
\centering
\begin{tabular}{lccc|cccc|cc}
\toprule
& \multicolumn{3}{c|}{\textbf{Image Level}} & \multicolumn{4}{c|}{\textbf{Pixel Level}} & \multicolumn{2}{c}{\textbf{Efficiency}} \\
\cmidrule(lr){2-4} \cmidrule(lr){5-8} \cmidrule(l){9-10}
\textbf{Model} &
\textbf{I-ROC} & \textbf{I-F1} & \textbf{I-AP} & 
\textbf{P-ROC} & \textbf{P-F1} & \textbf{P-AP} & \textbf{PRO} & 
\textbf{Mem [MB]} & \textbf{Inf [ms]} \\
\midrule
\multicolumn{10}{l}{\textit{MobileNet}} \\
\midrule
PaDiM      & 0.96 & 0.81 & 0.89 & \textbf{0.99} & \textbf{0.50} & 0.46 & \textbf{0.91} & 13.3 & 45 \\
CFA        & 0.91 & 0.74 & 0.75 & 0.97 & \textbf{0.50} & \textbf{0.47} & 0.82 & \textbf{6.8} & 18 \\
SSNet      & 0.69 & 0.46 & 0.48 & 0.72 & 0.04 & 0.02 & 0.60 & 7.6 & 126 \\
FastFlow   & 0.98 & 0.89 & 0.93 & 0.97 & 0.29 & 0.17 & 0.85 & 13.2 & 18 \\
RD4AD      & 0.82 & 0.59 & 0.53 & 0.79 & 0.05 & 0.02 & 0.64 & 11 & \textbf{15} \\
PatchCore  & \textbf{0.99} & \textbf{0.95} & \textbf{0.98} & 0.83 & 0.11 & 0.05 & 0.64 & 38.4 & 50 \\
STFPM      & 0.95 & 0.88 & 0.86 & 0.98 & 0.39 & 0.26 & 0.84 & 11.1 & 31 \\
\midrule
\multicolumn{10}{l}{\textit{WideResNet}} \\
\midrule
PaDiM      & 0.98 & 0.91 & 0.93 & \textbf{0.99} & 0.39 & 0.28 & 0.90 & 3900 & 24000 \\
CFA        & 0.96 & 0.86 & 0.91 & \textbf{0.99} & \textbf{0.66} & \textbf{0.68} & \textbf{0.94} & 148 & 195 \\
SSNet      & 0.61 & 0.72 & 0.64 & 0.70 & 0.08 & 0.03 & 0.62 & \textbf{143} & 649 \\
FastFlow   & 0.96 & 0.85 & 0.80 & 0.96 & 0.27 & 0.17 & 0.83 & 365 & 200 \\
RD4AD      & 0.98 & 0.88 & 0.90 & 0.97 & 0.24 & 0.16 & 0.91 & 645 & 240 \\
PatchCore  & \textbf{0.99} & \textbf{0.97} & \textbf{0.99} & 0.89 & 0.18 & 0.09 & 0.63 & 184.3 & 333 \\
STFPM      & 0.91 & 0.77 & 0.60 & \textbf{0.99} & 0.42 & 0.34 & 0.89 & 199.4 & \textbf{120} \\
\midrule
\multicolumn{10}{l}{\textit{DeiT Small}} \\
\midrule
PaDiM      & 0.99 & 0.91 & 0.97 & \textbf{0.99} & 0.56 & 0.46 & \textbf{0.93} & 205 & 1200 \\
CFA        & 0.72 & 0.48 & 0.47 & \textbf{0.99} & 0.53 & 0.47 & 0.85 & \textbf{91.3} & 32 \\
SSNet      & 0.58 & 0.38 & 0.37 & 0.67 & 0.03 & 0.01 & 0.56 & 96.9 & 32 \\
FastFlow   & \textbf{1.00} & \textbf{0.97} & \textbf{0.99} & 0.98 & 0.44 & 0.39 & 0.81 & 117 & \textbf{28} \\
RD4AD      & 0.94 & 0.82 & 0.84 & 0.96 & 0.28 & 0.17 & 0.79 & 156 & 103 \\
PatchCore  & 0.99 & 0.95 & \textbf{0.99} & 0.77 & 0.07 & 0.02 & 0.57 & 105.2 & 35 \\
STFPM      & 0.93 & 0.73 & 0.70 & \textbf{0.99} & 0.52 & 0.42 & 0.89 & 177 & 35 \\
Dinomaly   & \textbf{1.00} & 0.94 & \textbf{0.99} & \textbf{0.99} & \textbf{0.67} & \textbf{0.66} & 0.92 & 149 & 45 \\
\midrule
\multicolumn{10}{l}{\textit{DeiT Tiny}} \\
\midrule
PaDiM      & \textbf{0.99} & 0.89 & 0.96 & 0.99 & 0.51 & 0.38 & \textbf{0.92} & 52.8 & 150 \\
CFA        & 0.80 & 0.57 & 0.52 & 0.99 & 0.54 & 0.53 & 0.84 & \textbf{24.2} & \textbf{9} \\
SSNet      & 0.45 & 0.33 & 0.27 & 0.68 & 0.04 & 0.02 & 0.54 & 27.1 & 18 \\
FastFlow   & \textbf{0.99} & \textbf{0.93} & 0.94 & 0.96 & 0.26 & 0.19 & 0.79 & 30.5 & 14 \\
RD4AD      & 0.88 & 0.69 & 0.67 & 0.92 & 0.19 & 0.11 & 0.71 & 39.6 & 23 \\
PatchCore  & \textbf{0.99} & \textbf{0.93} & \textbf{0.98} & 0.78 & 0.10 & 0.04 & 0.64 & 31.4 & 19 \\
STFPM      & 0.84 & 0.60 & 0.54 & 0.98 & 0.44 & 0.30 & 0.85 & 47.5 & 14 \\
Tiny-Dinomaly & \textbf{0.99} & \textbf{0.93} & \textbf{0.98} & \textbf{1.00} & \textbf{0.69} & \textbf{0.66} & 0.90 & 37.6 & 19 \\
\bottomrule
\end{tabular}%
\caption{Anomaly detection benchmark results across models and backbones. Best results per backbone are in bold.  Memory (Mem) in MB; Inference Time (Inf) in ms.}
\label{tab:results}
\end{table*}

\section{Results}
\label{sec:results}

\subsection{Performance}
In this section, we discuss the performance of the eight VAD models using WideResNet and DeiT-Small backbones, as reported in Table \ref{tab:results} and in Figure \ref{fig:results-plot}, which displays a visual comparison between the models in terms of performance, memory requirements, and inference time.
\\
At the image level, most models achieve a very high I-ROC; instead, by examining the more challenging I-F1 score, which better reflects practical detection performance under class imbalance, a different picture emerges.
Some models such as FastFlow (0.97), PaDiM (0.91), PatchCore (0.95), and Dinomaly (0.94) stand out as the strongest performers with DeiT-Small. In contrast, STFPM (0.73) and SSNet (0.38) lag considerably behind, suggesting that despite acceptable ROC scores, these methods struggle to maintain a reliable performance at the image level.
\\
While many models show a generally strong ability to discriminate between normal and anomalous images, when considering pixel-level localization, measured by P-AP, the gap between methods becomes even more pronounced. Dinomaly (0.66) and CFA (0.68) achieve the best scores by a clear margin, respectively with DeiT and WideResNet, demonstrating superior precision in identifying anomalous regions. PaDiM (0.46) and STFPM (0.42) with Deit-Small follow at a distance, while the remaining methods struggle to produce well-localized anomaly maps.
\\
A particularly interesting observation concerns the effect of switching from CNN-based to Transformer-based backbones. Some methods benefit substantially from this transition: PaDiM improves its P-AP from 0.28 (WideResNet) to 0.46 (DeiT-Small), STFPM from 0.34 to 0.42 and FastFlow from 0.17 to 0.39. This suggests that representations produced by Vision Transformers encode richer spatial information that is advantageous for anomaly localization.
\\
An interesting insight is about PatchCore, one of the most famous and effective VAD models.
Despite near-perfect I-ROC (0.99) and I-F1 (0.95–0.97) scores, its pixel-level localization is consistently among the weakest, with P-AP of 0.09 and 0.02 for WideResNet and DeiT-Small respectively. This can be attributed to its memory bank design, which aggregates patch-level features without retaining spatial context. In road scenarios, where anomalies are localized objects embedded in a structured and spatially consistent environment, the absence of positional awareness becomes a critical limitation.

\subsection{Edge Deployment Analysis}
While most VAD models demonstrate strong performance when paired with DeiT-Small and WideResNet backbones, the practical feasibility of deploying these configurations directly onboard a vehicle remains a critical concern. The backbone models alone carry substantial memory footprints, 88.5 MB for DeiT-Small and 100 MB for WideResNet. Moreover, each method introduces additional components on top of the backbone: PatchCore, for instance, requires a memory bank, while RD4AD relies on a decoder. 
Therefore, backbone selection is a critical decision, as it can substantially drive up the total memory footprint and render many VAD models impractical for resource-constrained, onboard deployment.
\\
Given these constraints, we investigate two lightweight alternatives: MobileNet and  DeiT-Tiny.
The downstream effect on total system footprint is significant.
For example, PatchCore drops from 184 MB with WideResNet to 38.4 MB with MobileNet and also inference drops from 333 ms to just 50 ms per image, approaching the requirements of real-time processing (20 FPS).
\\
In terms of localization performance, CFA sees its P-AP drop from 0.68 with WideResNet to 0.47 with MobileNet. FastFlow also degrades noticeably when moving from DeiT-Small to DeiT-Tiny, with P-AP dropping from 0.39 to 0.19.
Tiny-Dinomaly stands out with a P-AP of 0.66, exactly matching its DeiT-Small counterpart, despite operating with a backbone roughly 4× smaller. This robustness likely reflects Dinomaly's reliance on self-supervised representations, which retain strong structural expressiveness even at a reduced scale.
\\
\\
From a system-level perspective, the proposed configurations are designed to be compatible with automotive-grade compute platforms. Models such as MobileNet and DeiT-Tiny, optimized for low-latency inference, are representative of workloads supported by platforms like the Qualcomm Snapdragon SA8295P \cite{lantronix_sa8295p_2026}, without requiring autonomy-grade hardware such as NVIDIA DRIVE Orin. Both MobileNet and DeiT-Tiny consistently achieve inference latencies below 50 ms, making them compatible with standard automotive sensing rates (20–30 FPS) and enabling near real-time anomaly detection.
\\
These results indicate that efficient onboard deployment is achievable for most VAD models without catastrophic performance degradation.
Among all evaluated models, Tiny-Dinomaly offers the best accuracy-efficiency trade-off, retaining competitive localization quality at substantially reduced memory cost.
Instead, when constraints about the memory are extremely heavy, PaDiM and CFA with MobileNet represent the most optimal choice.

\subsection{Anomaly Maps Evaluation}
As shown in Table \ref{tab:results}, Dinomaly achieves the highest pixel-level performance (P-F1 = 0.67, P-AP = 0.66) among all evaluated models, reflecting its strong ability to not only detect anomalies at the image level but also to precisely localize them within the scene. 
\\
This is further confirmed by qualitative inspection of the anomaly maps produced on the test set: as illustrated in Figures \ref{fig:heatmaps}a and \ref{fig:heatmaps}b, Dinomaly consistently generates well-focused activation regions that closely follow the spatial extent of the ground-truth anomaly, with limited spurious activations in normal areas of the scene.
\\
Despite this overall strong localization performance, a systematic examination of individual test images reveals two recurring failure modes.
First, anomalies that occupy a very small number of pixels in the image, such as distant or tiny objects, tend to be missed, making precise localization difficult (Figure \ref{fig:heatmaps}c). This behavior is consistent with the known limitations of patch-based feature representations, whose receptive field may be too coarse to capture fine-grained deviations at small scales. 
Second, scenes involving road curves introduce additional challenges: the perspective distortion can degrade the spatial coherence of the predicted anomaly map, leading to a wrong anomaly map (Figure \ref{fig:heatmaps}d). 
\\
These failure cases point to concrete open challenges for future work in AD4AD. 
Improving robustness to scale variation and addressing geometric distortions introduced by curved road geometries represent promising directions toward more reliable anomaly localization in real-world driving conditions.

\section{Conclusion}
\label{sec:conclusions}

This work presented the first systematic evaluation of Visual Anomaly Detection methods in the context of autonomous driving.
VAD was mainly designed and studied for industrial inspection, leaving road scenarios entirely unexplored.
We fill this gap by proposing a benchmark based on the AnoVox dataset that covers eight state-of-the-art methods.
\\
Our results demonstrated how VAD methods can transfer successfully to road scenes.
In particular, while most models achieve near-perfect image-level detection performance, localization is a harder task, and performance varies across models.
In addition, the choice of backbone has a significant impact on localization quality.
For example, we found out that replacing the WideResNet backbone, which is the standard in industrial anomaly detection, with a transformer architecture improves the performance in some models, including FastFlow, PaDiM, and STFPM.
\\
We also conducted an edge deployment analysis comparing lightweight architectures to evaluate the impact on performance, memory, and inference.
Specifically, we tested on the CNN-based MobileNet and on the transformer-based DeiT-Tiny.
The results indicate that efficient onboard deployment is achievable without catastrophic performance degradation.
Among all evaluated models, Tiny-Dinomaly offers the best accuracy-efficiency trade-off, retaining competitive localization quality at substantially reduced memory cost.
Instead, when constraints about the memory are extremely heavy, PaDiM and CFA with MobileNet represent the most optimal choice.
\\
Moving the discussion from the raw performance, this work highlights the practical value of producing anomaly maps with the goal of driver assistance.
By directing the attention of the driver toward the specific anomalous regions instead of generating generic alerts, VAD systems can support more informed decision-making in safety-critical situations.
\\
Future work directions should address the identified failure modes, namely the detection of small or distant objects and robustness to geometrically challenging scenes such as road curves, as well as explore multi-modal fusion with LiDAR and integration with downstream planning modules.

\bibliographystyle{ieeetr}
\bibliography{main}

\begin{thebibliography}{10}

\bibitem{draem}
V.~Zavrtanik, M.~Kristan, and D.~Sko{\v{c}}aj, ``Draem-a discriminatively trained reconstruction embedding for surface anomaly detection,'' in {\em Proceedings of the IEEE/CVF international conference on computer vision}, pp.~8330--8339, 2021.

\bibitem{diffusionad}
H.~Zhang, Z.~Wang, D.~Zeng, Z.~Wu, and Y.-G. Jiang, ``Diffusionad: Norm-guided one-step denoising diffusion for anomaly detection,'' 2025.

\bibitem{st_pyramid}
G.~Wang, S.~Han, E.~Ding, and D.~Huang, ``Student-teacher feature pyramid matching for anomaly detection,'' {\em arXiv:2103.04257}, 2021.

\bibitem{PaDiM}
T.~Defard, A.~Setkov, A.~Loesch, and R.~Audigier, ``{PaDiM}: A patch distribution modeling framework for anomaly detection and localization,'' in {\em Pattern Recognition. ICPR International Workshops and Challenges}, pp.~475--489, Springer International Publishing, 2021.

\bibitem{patch}
K.~Roth, L.~Pemula, J.~Zepeda, B.~Schölkopf, T.~Brox, and P.~Gehler, ``Towards total recall in industrial anomaly detection,'' {\em arXiv:2106.08265}, 2022.

\bibitem{lee2022cfa}
S.~Lee, S.~Lee, and B.~C. Song, ``Cfa: Coupled-hypersphere-based feature adaptation for target-oriented anomaly localization,'' {\em IEEE Access}, vol.~10, pp.~78446--78454, 2022.

\bibitem{yu2021fastflow}
J.~Yu, Y.~Zheng, X.~Wang, W.~Li, Y.~Wu, R.~Zhao, and L.~Wu, ``Fastflow: Unsupervised anomaly detection and localization via 2d normalizing flows,'' 2021.

\bibitem{Bogdoll_2023_adbenchmarksurvey}
D.~Bogdoll, S.~Uhlemeyer, K.~Kowol, and J.~M. Zöllner, ``Perception datasets for anomaly detection in autonomous driving: A survey,'' in {\em 2023 IEEE Intelligent Vehicles Symposium (IV)}, p.~1–8, IEEE, June 2023.

\bibitem{chan2021segmentmeifyoucanbenchmarkanomalysegmentation}
R.~Chan, K.~Lis, S.~Uhlemeyer, H.~Blum, S.~Honari, R.~Siegwart, P.~Fua, M.~Salzmann, and M.~Rottmann, ``Segmentmeifyoucan: A benchmark for anomaly segmentation,'' 2021.

\bibitem{hendrycks2022scalingoutofdistributiondetectionrealworld}
D.~Hendrycks, S.~Basart, M.~Mazeika, A.~Zou, J.~Kwon, M.~Mostajabi, J.~Steinhardt, and D.~Song, ``Scaling out-of-distribution detection for real-world settings,'' 2022.

\bibitem{lostandfound}
P.~Pinggera, S.~Ramos, S.~Gehrig, U.~Franke, C.~Rother, and R.~Mester, ``Lost and found: detecting small road hazards for self-driving vehicles,'' in {\em International Conference on Intelligent Robots and Systems}, pp.~1099--1106, 10 2016.

\bibitem{fishyscapes}
H.~Blum, P.-E. Sarlin, J.~Nieto, R.~Siegwart, and C.~Cadena, ``The fishyscapes benchmark: Measuring blind spots in semantic segmentation,'' {\em International Journal of Computer Vision}, vol.~129, pp.~1--17, 11 2021.

\bibitem{bogdoll2024anovox}
D.~Bogdoll, I.~Hamdard, L.~N. R{\"o}{\ss}ler, F.~Geisler, M.~Bayram, F.~Wang, J.~Imhof, M.~De~Campos, A.~Tabarov, Y.~Yang, {\em et~al.}, ``Anovox: A benchmark for multimodal anomaly detection in autonomous driving,'' in {\em European Conference on Computer Vision}, pp.~206--223, Springer, 2024.

\bibitem{bogdoll2022anomaly}
D.~Bogdoll, M.~Nitsche, and J.~M. Z{\"o}llner, ``Anomaly detection in autonomous driving: A survey,'' in {\em Proceedings of the IEEE/CVF conference on computer vision and pattern recognition}, pp.~4488--4499, 2022.

\bibitem{barusco2024paste}
M.~Barusco, F.~Borsatti, D.~D. Pezze, F.~Paissan, E.~Farella, and G.~A. Susto, ``Paste: Improving the efficiency of visual anomaly detection at the edge,'' {\em arXiv preprint arXiv:2410.11591}, 2024.

\bibitem{rd4ad}
H.~Deng and X.~Li, ``Anomaly detection via reverse distillation from one-class embedding,'' 2022.

\bibitem{rolih2025supersimplenet}
B.~Rolih, M.~Fu{\v{c}}ka, and D.~Sko{\v{c}}aj, ``Supersimplenet: Unifying unsupervised and supervised learning for fast and reliable surface defect detection,'' in {\em International Conference on Pattern Recognition}, pp.~47--65, Springer, 2025.

\bibitem{guo2025dinomaly}
J.~Guo, S.~Lu, W.~Zhang, F.~Chen, H.~Li, and H.~Liao, ``Dinomaly: The less is more philosophy in multi-class unsupervised anomaly detection,'' in {\em Proceedings of the Computer Vision and Pattern Recognition Conference}, pp.~20405--20415, 2025.

\bibitem{touvron2021training}
H.~Touvron, M.~Cord, M.~Douze, F.~Massa, A.~Sablayrolles, and H.~J{\'e}gou, ``Training data-efficient image transformers \& distillation through attention,'' in {\em International conference on machine learning}, pp.~10347--10357, PMLR, 2021.

\bibitem{barusco2026continual}
M.~Barusco, F.~Borsatti, D.~Petrovic, D.~D. Pezze, and G.~A. Susto, ``Continual visual anomaly detection on the edge: Benchmark and efficient solutions,'' {\em arXiv preprint arXiv:2604.06435}, 2026.

\bibitem{lantronix_sa8295p_2026}
{Lantronix, Inc.}, ``{SA8295P} automotive development platform.'' Lantronix Product Page, 2026.

\end{thebibliography}

\end{document}